\documentclass[sigconf]{acmart}

\AtBeginDocument{
  }

\setcopyright{acmcopyright}
\copyrightyear{2024}
\acmYear{2024}
\acmDOI{XXXXXXX.XXXXXXX}

\acmConference[KDD-UC '24]{KDD Undergraduate Consortium}{August 25--29, 2024}{Barcelona, Spain}

\acmISBN{978-1-4503-XXXX-X/24/08}

\usepackage{subfig}           
\usepackage{overpic}
\usepackage{booktabs}
\usepackage{caption}
\usepackage{geometry}
\usepackage{tabularx}
\usepackage{xcolor}
\usepackage{graphicx}
\usepackage{array}
\usepackage{makecell}
\usepackage{balance}
\begin{document}

\title{Deep Imbalanced Regression to Estimate Vascular Age from PPG Data: a Novel Digital Biomarker for Cardiovascular Health}

\author{Guangkun Nie$^{1,2}$, Qinghao Zhao$^3$, Gongzheng Tang$^{1,2}$, Jun Li$^4$, Shenda Hong\texorpdfstring{\textsuperscript{*}}{}$^{1,2}$}

\affiliation{%
  \institution{$^1$ Institute of Medical Technology, Health Science Center of Peking University, Beijing, China}
  \country{}
}

\affiliation{%
  \institution{$^2$ National Institute of Health Data Science, Peking University, Beijing, China}
  \country{}
}

\affiliation{%
  \institution{$^3$ Department of Cardiology, Peking University People’s Hospital, Beijing, China}
  \country{}
}

\affiliation{%
  \institution{$^4$ College of Electronic Science and Engineering, Jilin University, Changchun, China}
  \country{}
}

\email{
nieguangkun@stu.pku.edu.cn,
qhzhao@hsc.pku.edu.cn,
gztang@hsc.pku.edu.cn}

\email{lijun2020@mails.jlu.edu.cn, 
hongshenda@pku.edu.cn}

\begin{abstract} 
Photoplethysmography (PPG) is emerging as a crucial tool for monitoring human hemodynamics, with recent studies highlighting its potential in assessing vascular aging through deep learning. However, real-world age distributions are often imbalanced, posing significant challenges for deep learning models. In this paper, we introduce a novel, simple, and effective loss function named the Dist Loss to address deep imbalanced regression tasks. We trained a one-dimensional convolutional neural network (Net1D) incorporating the Dist Loss on the extensive UK Biobank dataset (n=502,389) to estimate vascular age from PPG signals and validate its efficacy in characterizing cardiovascular health. The model's performance was validated on a 40\% held-out test set, achieving state-of-the-art results, especially in regions with small sample sizes. Furthermore, we divided the population into three subgroups based on the difference between predicted vascular age and chronological age: less than -10 years, between -10 and 10 years, and greater than 10 years. We analyzed the relationship between predicted vascular age and several cardiovascular events over a follow-up period of up to 10 years, including death, coronary heart disease, and heart failure. Our results indicate that the predicted vascular age has significant potential to reflect an individual's cardiovascular health status. Our code is available at \url{https://github.com/Ngk03/AI-vascular-age}.
\end{abstract}


\maketitle

\section{Introduction}
Photoplethysmography (PPG) is a widely utilized biosignal for assessing human hemodynamics, offering advantages such as convenience, non-invasiveness, and continuous monitoring. Recent advancements in deep learning technology have further unlocked the potential of PPG in managing individual cardiovascular health, demonstrating promising progress \cite{franklin2023synchronized, edgar2024automated, nie2024review}. On this basis, several studies have shown the value of PPG in evaluating vascular aging \cite{dall2020prediction,shin2022photoplethysmogram,ferizoli2024arterial,chen2024predicting}, a crucial indicator of cardiovascular health. In this paper, we use PPG along with deep learning to predict vascular age, providing a refined representation of cardiovascular health levels.

However, in the real world, the age distribution of individuals is often imbalanced, posing significant challenges for deep learning models. When trained on imbalanced samples, models tend to predict towards regions with higher sample density to reduce overall prediction error, leading to significant biases in regions with fewer samples \cite{yang2021delving, ren2022balanced}. This approach results in suboptimal parameters because the model fails to learn robust and meaningful features for prediction, rendering the predictions unreliable.

\begin{figure}
    \centering
    \includegraphics[width=\linewidth]{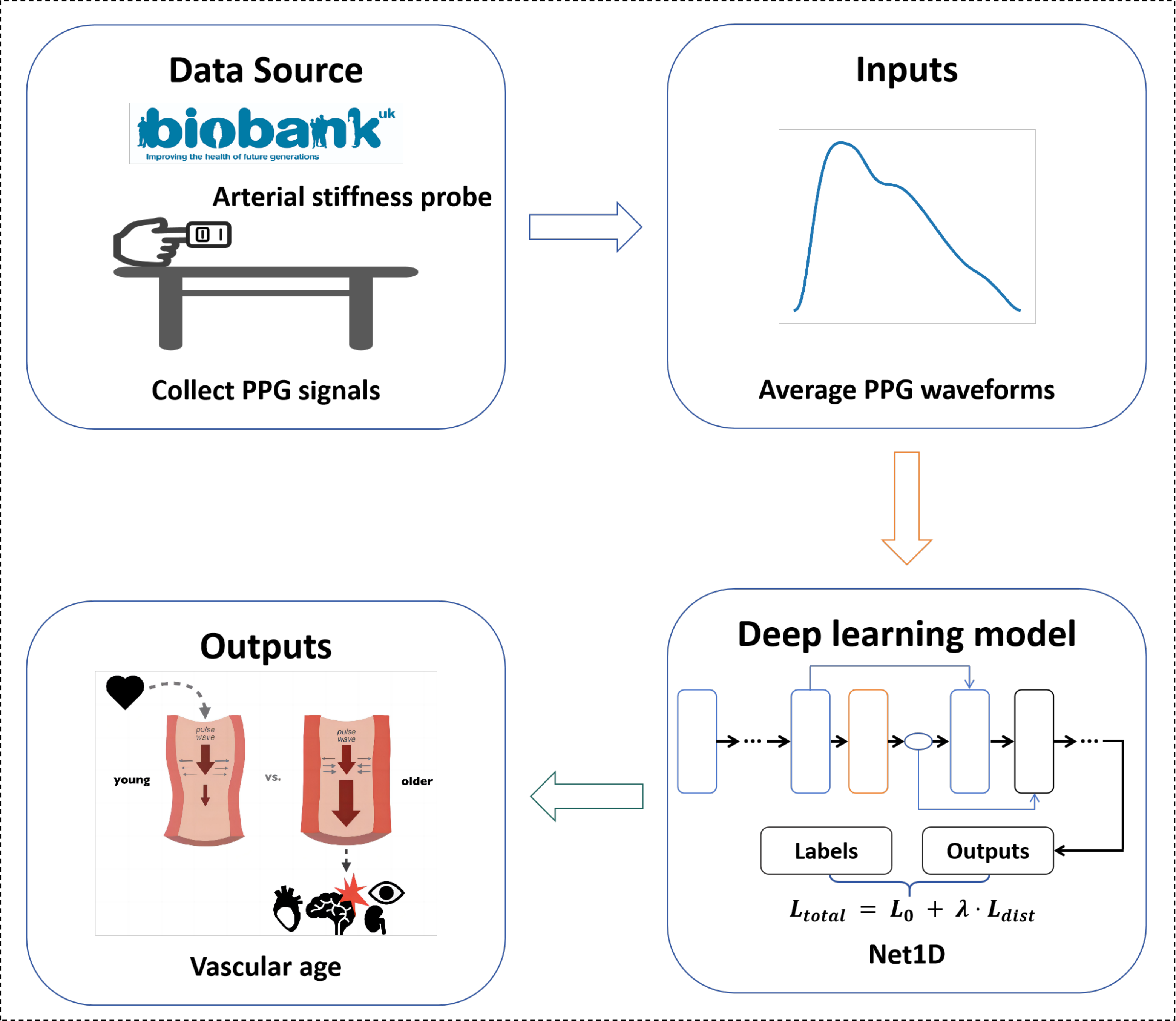}
    \caption{Workflow of our vascular age estimation method.}
    \label{overview}
\end{figure}

To address this issue, we introduce a novel loss function based on data distribution priors, called the Dist loss. This loss function encourages the model to align predictions more closely with the desired label distribution, reducing the tendency to over-predict in high-density regions. Additionally, we validated the effectiveness of our proposed approach using a large cohort database, the UK BioBank (UKBB) (n=502,389). Our results demonstrate significantly improved performance and highlight the potential of vascular age in representing individual cardiovascular health levels. An overview of the workflow for vascular age estimation is shown in Figure \ref{overview}.

Overall, the contributions of this paper are as follows:

\begin{itemize}
    \item We propose a novel, simple, and effective loss function, the Dist loss, to effectively mitigate prediction biases in deep imbalanced regression tasks. This loss function significantly improves the performance of the model in regions with few samples, achieving state-of-the-art (SOTA) results.
    
    \item Utilizing the large cohort database UKBB, we analyze the relationship between predicted vascular age and certain outcomes (e.g., death, coronary heart disease (CHD), heart failure) in subjects. This analysis validates the application value of vascular age predicted by deep learning models in representing individual cardiovascular health levels.
\end{itemize}

\section{Methods}
\subsection{Problem definition}
For a given PPG dataset \( D = \{X, Y\} \): \( X = \{X_{1}, X_{2}, \dots, X_{N}\} \in \mathbb{R}^{N \times L} \) represents the PPG signals of \( N \) instances, each with a length of \( L \), and \( Y = \{Y_{1}, Y_{2}, \dots, Y_{N}\} \in \mathbb{R}^{N} \) represents the corresponding labels. The label space, which includes all possible label values, is \( y = \{y_{1}, y_{2}, \dots, y_{n}\} \in \mathbb{R}^{n} \), sorted in ascending/descending order. The goal of model training is to find the optimal parameters that accurately capture the relationship between the PPG signals \( X \) and the labels \( Y \). This is achieved by minimizing a loss function that measures the discrepancy between the model's predictions and the actual labels. Mean squared error (MSE) and mean absolute error (MAE) are commonly used for this purpose in regression tasks.

When chronological ages are used as labels, the trained model predicts an individual's chronological age based on their PPG signal. For healthy individuals, the predicted age should align closely with their actual chronological age. Conversely, individuals with abnormal hemodynamics, commonly associated with cardiovascular diseases, will have PPG signals resembling those of older individuals. On the other hand, very healthy individuals will exhibit PPG signals similar to those of younger individuals. Therefore, the model's predictions reflect an individual's cardiovascular health status, which we term as vascular age.

\subsection{Model architecture}
The model architecture utilized to predict vascular age is Net1D \cite{hong2020holmes}, a one-dimensional convolutional neural network (CNN). This model incorporates sequential convolutional blocks with residual connections \cite{he2016deep} and squeeze-and-excitation modules \cite{hu2018squeeze}. Leveraging the proven efficacy of CNNs in diverse PPG-based tasks \cite{osathitporn2023rrwavenet, torres2020multi}, the model effectively captures nuanced and distinctive features from PPG signals associated with vascular aging.

\subsection{Deep imbalanced regression}
\paragraph{Overview}
Imbalanced data distributions pose significant challenges for deep learning models, hindering their ability to extract meaningful and generalizable features. In biomedical applications, such as estimating vascular age from PPG data, this issue becomes even more pronounced due to the critical need for accurate and reliable predictions in healthcare.

\begin{figure}
\centering
\begin{minipage}[b]{0.495\linewidth}
  \centering
  \subfloat[]{\includegraphics[width=\linewidth]{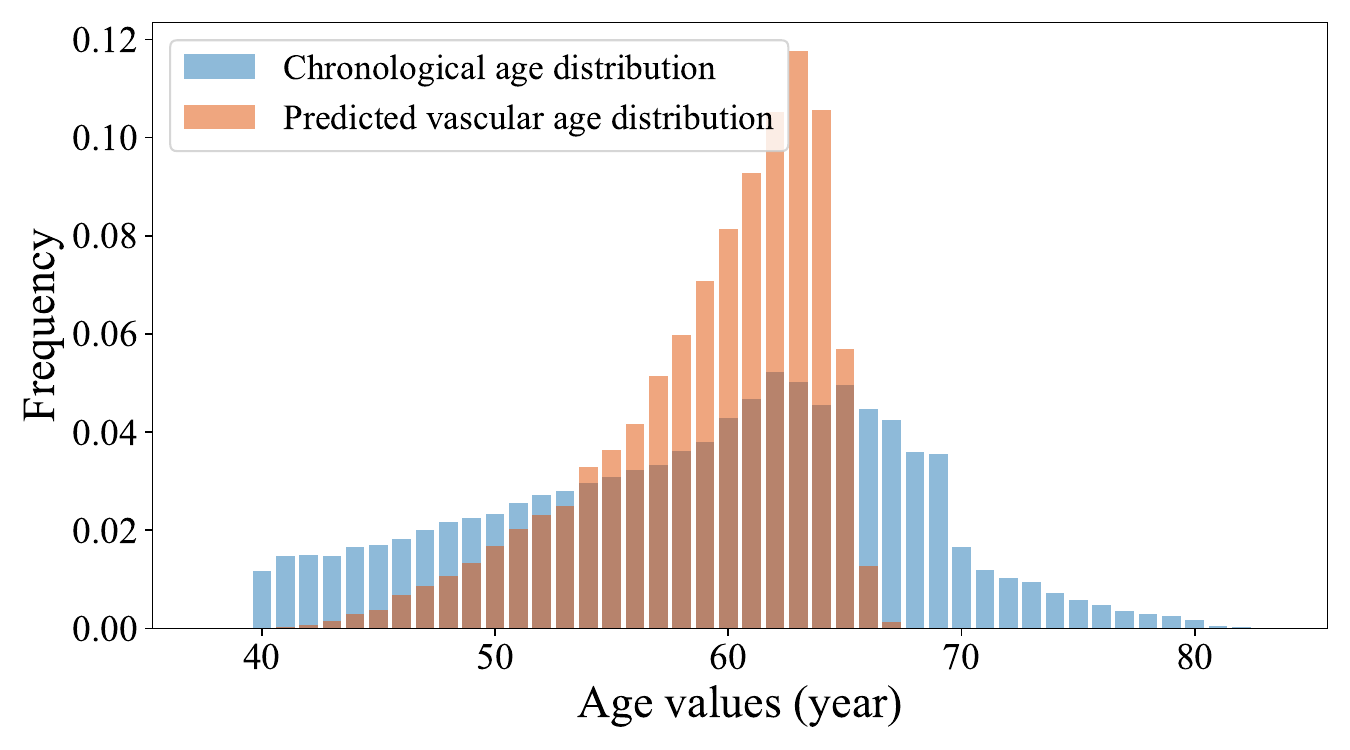}\label{fig:plain age distribution}}
\end{minipage}
\hfill
\begin{minipage}[b]{0.495\linewidth}
  \centering
  \subfloat[]{\includegraphics[width=\linewidth]{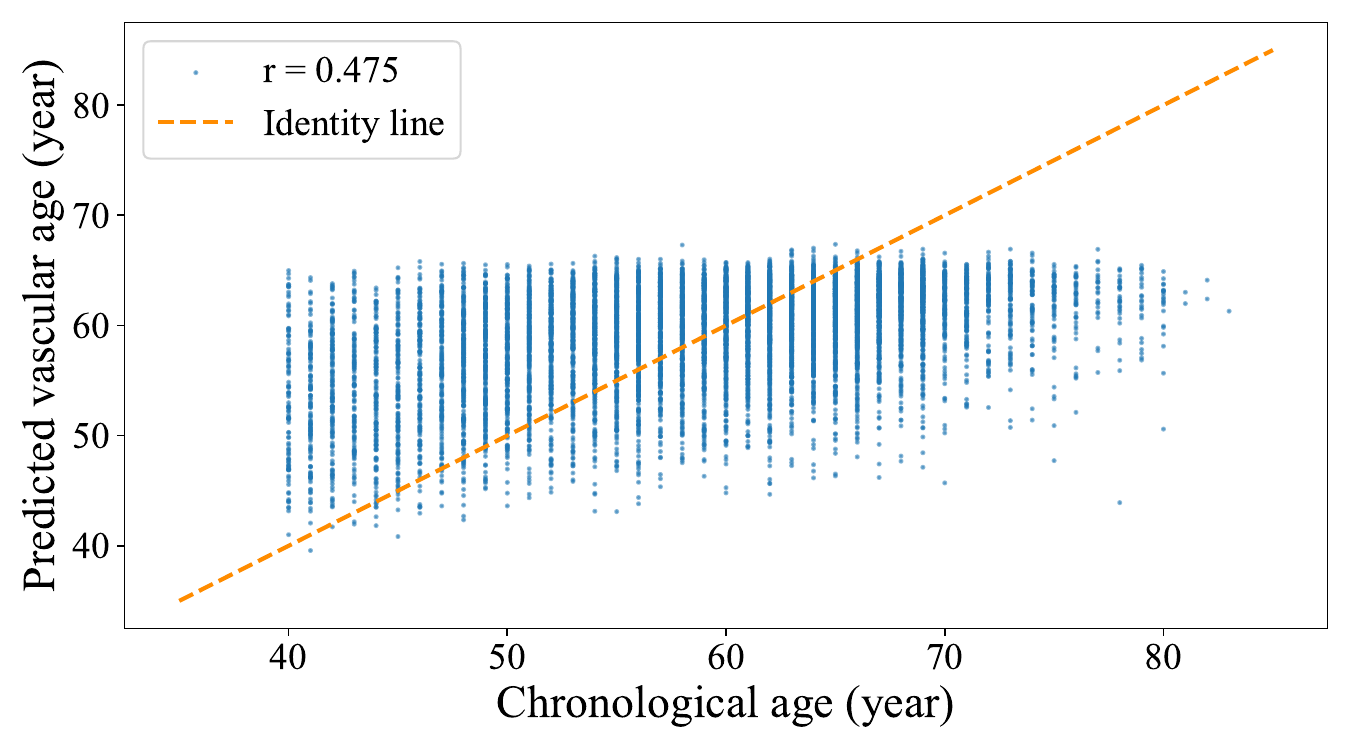}\label{fig:plain scatter}}
\end{minipage}

\vspace{0em} 

\begin{minipage}[b]{0.495\linewidth}
  \centering
  \subfloat[]{\includegraphics[width=\linewidth]{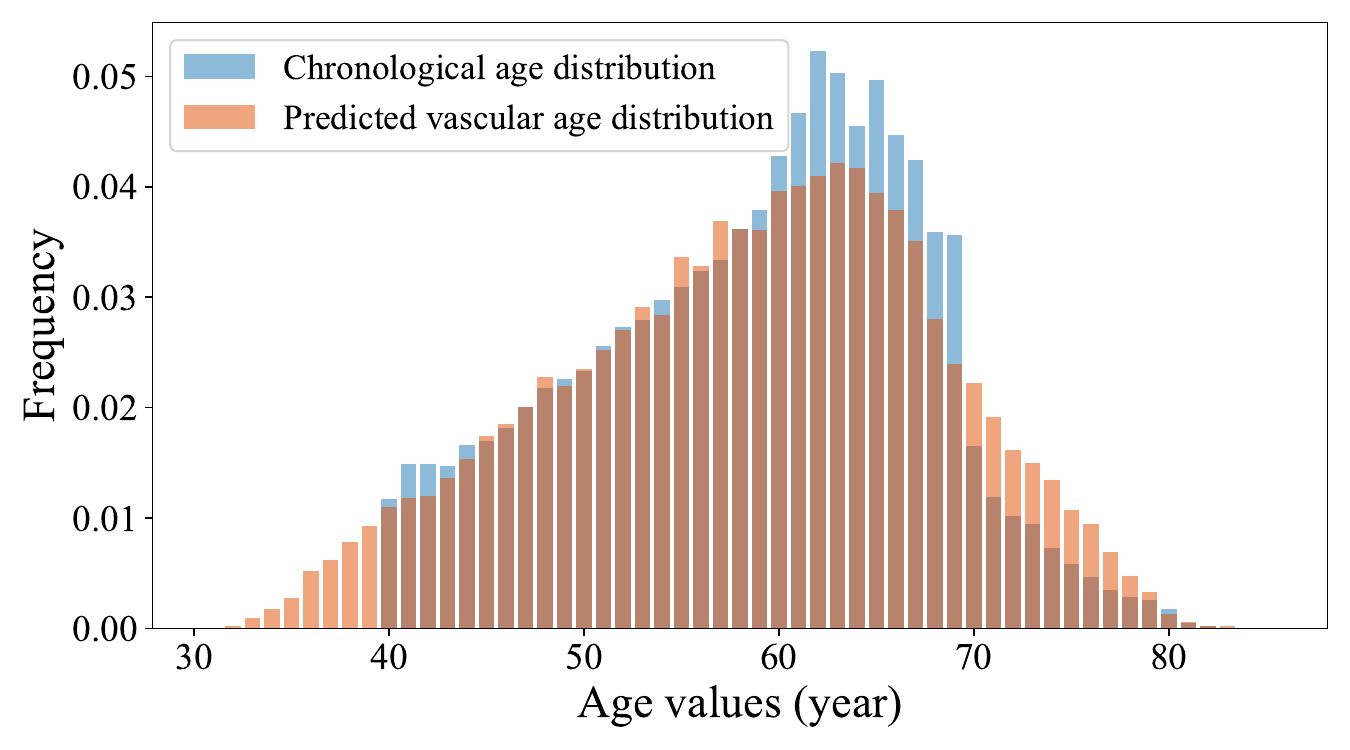}\label{fig: dist_based age distribution}}
\end{minipage}
\hfill
\begin{minipage}[b]{0.495\linewidth}
  \centering
  \subfloat[]{\includegraphics[width=\linewidth]{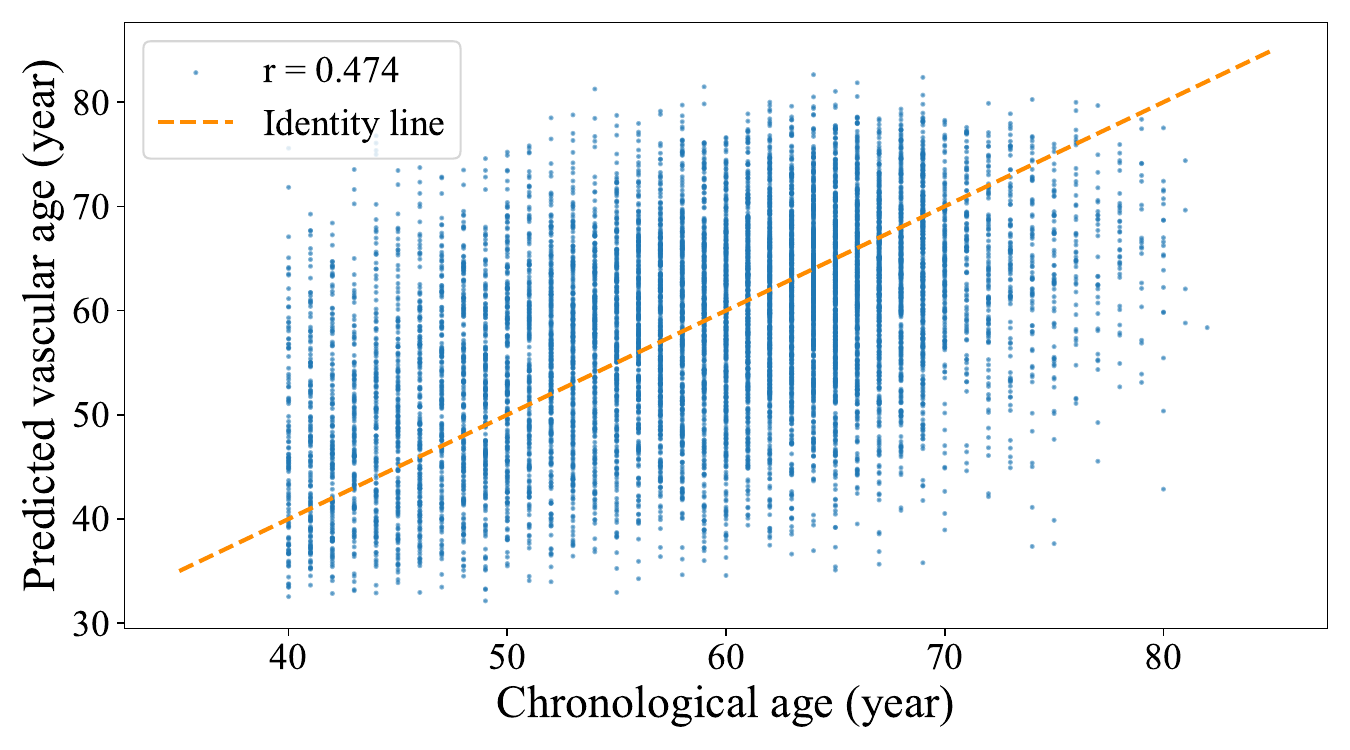}\label{fig:dist_based scatter result}}
\end{minipage}
\caption{Comparison between the predicted vascular age by the plain model (a, b) and our method (c, d), $r$ denotes Pearson correlation coefficient. (a) Illustrates the distribution of chronological age (light blue bars) and the corresponding vascular age predicted by the plain model (orange bars) in the UKBB dataset. (b) Presents a scatter plot highlighting the disparity between chronological age and predicted vascular age by the plain model. (c) Depicts the distribution of chronological age (light blue bars) and the corresponding vascular age predicted by our method (orange bars). (d) Displays a scatter plot showcasing the higher consistency between chronological age and predicted vascular age achieved by our method.}
\label{results demo}
\end{figure}

As shown in Figure \ref{fig:plain age distribution}, the age distribution within the UKBB dataset demonstrates a noticeable skew, with a concentration of samples around the mean age and fewer samples at the extremes, especially among older individuals. This imbalance leads to a tendency to over-predict in high-density regions and introduces significant biases in regions with fewer samples, rendering the model predictions unreliable \cite{yang2021delving, ren2022balanced, gong2022ranksim}.

Models trained solely on MAE or MSE loss, referred to as "plain models," often perform suboptimally on imbalanced data distributions. For demonstration purposes, the "plain models" discussed below are trained using the MAE loss function, although similar behavior was observed with MSE. As shown in Figure \ref{fig:plain age distribution}, the plain model's predictions of vascular age distribution on the UKBB dataset tend to be biased towards regions with higher sample density. This results in inaccurate representations of the vascular age distribution, despite the plain model achieving satisfactory performance on conventional evaluation metrics such as MAE and Root Mean Squared Error (RMSE). Further analysis in Figure \ref{fig:plain scatter} underscores the discrepancy between the chronological age distribution and the predicted vascular age distribution generated by the plain model. Notably, the predicted vascular age is unevenly distributed across age ranges, with younger individuals often predicted as older and vice versa, highlighting how imbalanced data distribution hampers the model's ability to acquire optimal features.

To tackle the challenges arising from data imbalance, we introduce a simple yet effective solution termed as \textbf{the Dist loss}. This novel loss function harnesses prior knowledge of label distribution in the training data to aid the model in learning useful and meaningful features, thus mitigating prediction bias.

\paragraph{\textbf{The Dist loss}}
As previously noted, a significant disparity exists between the distribution of labels and the model predictions, primarily caused by the data imbalance. Intuitively, one possible solution involves minimizing the gap between these two distributions. To address this, we propose a novel loss function called the Dist loss, specifically designed to align model predictions more closely with the desired label distribution.

To evaluate the distance between two distributions, commonly used metrics include Kullback-Leibler (KL) divergence and Jensen-Shannon (JS) divergence. Although the computation of these metrics is straightforward in classification tasks, it poses challenges in regression tasks due to the potentially infinite number of possible label values. Additionally, existing methods like kernel density estimation (KDE) \cite{yang2021delving} for label distribution estimation are non-differentiable, hindering gradient backpropagation and thereby preventing model parameter updates. Hence, instead of directly estimating the distribution of predictions, we introduce a surrogate approach.

For the given PPG dataset \( D = \{X, Y\} \) comprising \( N \) instances, as described in Section 3.1, the corresponding model predictions are denoted as \( \hat{Y} \). To calculate the Dist loss, we first estimate the label distribution using KDE, resulting in the probability distribution \( P(y) \) for all possible label values \( y = (y_{1}, y_{2}, \dots, y_{n}) \), sorted in ascending/descending order. For continuous labels with an infinite number of possible values, discretization is applied. We then compute the expected occurrence numbers \( N^{E} = \{N_{1}, N_{2}, \dots, N_{n}\} \) for each possible label, calculated as \( \left \lfloor N \cdot P(y) \right \rfloor \), where \( \left \lfloor \cdot \right \rfloor \) denotes the floor function. Based on this, we generate the 'expected labels' \( Y^{E} \) for \( N \) instances, as shown in Equation \ref{exp labels}.

\begin{equation}
    \boxed{Y^{E} = (\underbrace{y_{1}, \dots, y_{1}}_{N_1}, \dots, \underbrace{y_{i}, \dots, y_{i}}_{N_i},\dots,\underbrace{y_{n}, \dots, y_{n}}_{N_n})}
    \label{exp labels}
\end{equation}

Next, we sort the model predictions $\hat{Y}$ in the same order as $y$, resulting in the sorted output $\hat{Y}_{sort}$. Finally, we compute the Dist loss $L_{dist}$ using Equation \ref{dist loss}, where $M(\cdot)$ represents a function that measures the distance between two variables, such as MAE or MSE. The only seemingly non-differentiable part of this process is the sorting function, which can be addressed using the fast differentiable sorting method proposed by Blondel et al. \cite{blondel2020fast}. During training, the total loss function is defined in Equation \ref{loss}, where $\lambda$ is a hyperparameter that adjusts the importance of distribution similarity.

\begin{equation}
    \boxed{L_{dist} = M(\hat{Y}_{sort}, Y_{sort}^{E})} 
    \label{dist loss}
\end{equation}

\begin{equation}
    \boxed{L_{total} = M(\hat{Y}, Y) + \lambda \cdot L_{dist}}
    \label{loss}
\end{equation}

\section{Experiments and Results} 
\subsection{Experimental setup}
\subsubsection{Dataset and preprocessing}
This study utilizes data from the UK Biobank (UKBB) dataset, which includes genetic and health information from over 500,000 individuals across the United Kingdom. PPG data were extracted from UKBB field 4205. The data were collected using the PulseTrace PCA2 device, which captured average waveforms of 10-15 second PPG signals standardized to a length of 100. We randomly divided the 502,389 participants into training (60\%) and test (40\%) sets, with each participant having 0-4 PPG records. This division resulted in 143,196 records (n=126,216) for training and 100,172 records (n=86,027) for testing. Each PPG signal was then normalized using Equation \ref{normalization}, where $mean(\cdot)$ and $std(\cdot)$ denote the mean and standard deviation functions, respectively.

\begin{equation}
    normalization(X_{i}) = \frac{X_{i} - mean(X_{i})}{std(X_{i})}
    \label{normalization}
\end{equation}

\subsubsection{Evaluation metrics}
To evaluate and compare our method with existing approaches for deep imbalanced regression, we utilize a comprehensive set of metrics beyond the conventional Pearson correlation coefficient, MAE, and RMSE. These conventional metrics are insufficient for accurately assessing model performance in highly imbalanced data scenarios. For the labels $Y = (Y_1,Y_2,\dots,Y_N)$, their occurrence probabilities $P_{Y} = (P_{Y1},P_{Y2},\dots,P_{YN})$, and the corresponding predictions $\hat{Y} = (\hat{Y}_1, \hat{Y}_2, \dots, \hat{Y}_N)$, the formulas for these evaluation metrics are as follows:

\paragraph{Pearson Correlation Coefficient}
The Pearson correlation coefficient measures the linear correlation between two variables, ranging from -1 to 1. A value of 1 indicates a perfect positive linear relationship, -1 indicates a perfect negative linear relationship, and 0 indicates no linear relationship. The formula for the Pearson correlation coefficient $r$ is given by Equation \ref{Coef}, where $\bar{Y}$ and $\bar{\hat{Y}}$ are the means of the labels $Y$ and the predictions $\hat{Y}$, respectively.

\begin{equation}
r = \frac{\sum_{i=1}^{N} (Y_i - \bar{Y})(\hat{Y}_i - \bar{\hat{Y}})}{\sqrt{\sum_{i=1}^{N} (Y_i - \bar{Y})^2 \sum_{i=1}^{N} (\hat{Y}_i - \bar{\hat{Y}})^2}}
\label{Coef}
\end{equation}

\paragraph{MAE \& RMSE} MAE and RMSE are commonly used metrics to evaluate model performance in regression tasks. The formulas for MAE and RMSE are shown in Equations \ref{MAE} and \ref{RMSE}, respectively.

\begin{equation}
    \text{MAE} = \frac{1}{N} \sum_{i=1}^{N} \lvert Y_{i} - \hat{Y}_{i} \rvert
    \label{MAE}
\end{equation}

\begin{equation}
    \text{RMSE} = \sqrt[]{\frac{1}{N} \sum_{i=1}^{N} (Y_i - \hat{Y}_{i})^2}
    \label{RMSE}
\end{equation}

\paragraph{Weighted MAE \& RMSE} Standard MAE and RMSE are insufficient for evaluating model performance with highly imbalanced labels. To address this, we use weighted MAE and RMSE for a more accurate assessment. The formulas for weighted MAE and RMSE are shown in Equations \ref{WMAE} and \ref{WRMSE}.

\begin{equation}
    \text{Weighted MAE} = \frac{1}{N} \sum_{i=1}^{N} \left [\lvert Y_{i} - \hat{Y}_{i} \rvert \cdot \frac{P_{Yi}}{\text{mean}(P_Y)} \right ]
    \label{WMAE}
\end{equation}

\begin{equation}
    \text{Weighted RMSE} = \sqrt[]{\frac{1}{N} \sum_{i=1}^{N} \left [(Y_i - \hat{Y}_{i}) \cdot \frac{P_{Yi}}{\text{mean}(P_Y)}\right ]^2} 
    \label{WRMSE}
\end{equation}

\paragraph{Overlap ratio}
The overlap ratio (OR) is a metric used to assess the similarity between two distributions by calculating the ratio of the area of overlap to the total area under the distributions. The OR ranges from 0 to 1, where a value of 1 indicates perfect overlap and 0 indicates no overlap. The formula for OR is presented in Equation \ref{OR}, where $O_{y} = \{O_1, O_2, \dots, O_n\}$ and $\hat{O}_{y} = \{\hat{O}_1, \hat{O}_2, \dots, \hat{O}_n\}$ represent the observed numbers in labels and model predictions for each possible label value in the label space $y = \{y_1, y_2, \dots, y_n\}$.

\begin{equation}
    \text{OR} = \frac{\sum_{y_1}^{y_n} min(O_i,\hat{O}_i)}{\sum_{y_1}^{y_n} max(O_i,\hat{O}_i)} 
    \label{OR}
\end{equation}

\subsubsection{Compared methods}
To evaluate the effectiveness of the Dist loss, we conducted comparisons with several existing methods designed for deep imbalanced regression.

\paragraph{LDS \& FDS} LDS (label distribution smoothing) and FDS (feature distribution smoothing) are two modules proposed in \cite{yang2021delving}. These modules are designed to calibrate label and learned feature distributions, respectively.

\paragraph{Ranksim} Ranksim, proposed in \cite{gong2022ranksim}, leverages label similarity to enhance model performance. It ensures that samples with similar labels are positioned closer together in the feature space, while those with dissimilar labels are positioned further apart.

\paragraph{Balanced MSE} Balanced MSE \cite{ren2022balanced} is a modified loss function derived from the standard MSE. It is designed to restore balanced predictions by leveraging the training label distribution prior to making a statistical conversion.

\subsubsection{Statistical analysis}
While the Dist loss serves as a general method for mitigating model bias caused by data imbalance, some bias, although significantly reduced, remains inevitable. Therefore, inspired by \cite{le2022using}, we conducted a linear regression on the residuals as a function of age for the model and utilized it to correct each prediction during clinical validation, taking the corrected vascular age as the final predictions. It is noteworthy that experiments showed that without using the Dist loss during model training, the corrected vascular age exhibited little correlation with individuals' cardiovascular health. This was primarily because the features extracted by the plain model were inaccurate and meaningless, highlighting the importance of the Dist loss in enabling the model to learn accurate and meaningful vascular age representations.

During clinical validation, to evaluate the effectiveness of the predicted vascular age, we categorized the samples into three subgroups based on the difference between corrected vascular age and chronological age: less than -10 years, between -10 and 10 years, and greater than 10 years. For each group, we summarized the mean and standard deviation for continuous variables, including age and body mass index (BMI), and provided counts and percentages for categorical variables such as ethnic background and sex. We also examined the statistical differences between these three groups using Student's t-test and the Kruskal-Wallis H test for continuous variables and the chi-square test for categorical variables. Additionally, we applied the Cochran-Armitage Trend Test to examine baseline trends across the subgroups. During the analysis, subjects lost to follow-up (n=153) were excluded.

For cardiovascular events analysis, we employed the Cox proportional hazards regression model to report hazard ratios (HR) and 95\% confidence intervals (95\% CI). A follow-up period of up to 10 years was set for this analysis. The analysis involved three levels of adjustments: (1) age, sex, and ethnic background; (2) age, sex, ethnic background, hypertension, and diabetes; and (3) age, sex, ethnic background, hypertension, diabetes, BMI, and dyslipidemia. The proportional hazards assumption was verified using Schoenfeld residuals. Additionally, we demonstrated the adjusted survival curves from the Cox models, stratified by age, sex, and ethnic background.

\subsubsection{Implementation Details} All methods were integrated with Net1D for equitable comparisons. In our method, we used MAE as $M(\cdot)$ to quantify the discrepancy between two variables, with $\lambda$ set to 1, indicating that distribution similarity is equally important as total errors. The training process for all methods employed a batch size of 2048, 50 epochs, and a learning rate of 3e-3. The Adam optimizer with cosine learning rate decay was utilized, and a weight decay of 1e-4 was applied to prevent overfitting. Additionally, based on the probability of label occurrence, we identified the region with label occurrence probabilities less than one-third of the maximum as \textbf{the few-shot region}.

\subsection{Main results}
The performance of all methods regarding the Pearson correlation coefficient is summarized in Table \ref{overall performance}, while the results in the few-shot region for all methods are presented in Table \ref{tab:performance_metrics}, with the top two results for each metric highlighted in bold and red. Additionally, the predictions of the Dist loss-based model are demonstrated in Figure \ref{results demo}. It can be observed that the distribution of the predicted vascular age closely matches the chronological age, and our method outperforms all compared methods in the few-shot region, significantly reducing prediction bias, while still maintaining good performance in terms of the overall Pearson correlation coefficient.

\begin{table}
\centering
\caption{Comparison of Pearson correlation coefficients among different methods for predicting vascular age.}
\begin{tabular}{lc}
\\ \hline
               & \textbf{Pearson correlation coefficient} \\ \hline
Plain          & \textbf{\textcolor{red}{0.475}}                         \\
+ LDS          & 0.440                           \\
+ FDS          & 0.448                           \\
+ LDS \& FDS   & 0.434                           \\
+ Ranksim      & 0.392                           \\
+ Balanced MSE & 0.422                           \\
+ \textbf{Ours}         & \textbf{\textcolor{red}{0.474}}                           \\ \hline
\end{tabular}
\label{overall performance}
\end{table}

\begin{table*}
    \centering
    \caption{Results of different methods in the few-shot region, where $r$ represents Pearson correlation coefficient.}
    \resizebox{\textwidth}{!}{%
    \begin{tabular}{lcccccccc}
        \toprule
        \multicolumn{9}{c}{\textbf{Few-shot region}} \\
        \midrule
        & \textbf{$r$} & \textbf{MAE} & \textbf{RMSE} & \textbf{Overlap ratio} & \textbf{Weighted MAE} & \textbf{Weighted RMSE} & \textbf{MAE / Overlap ratio} & \textbf{RMSE / Overlap ratio} \\
        \midrule
        Plain        & 0.637 & 12.820 & 13.786 & 0.015 & 6.522 & 7.524 & 865.377 & 930.543 \\
        + LDS    & \textbf{\textcolor{red}{0.683}} & \textbf{\textcolor{red}{10.585}} & \textbf{\textcolor{red}{12.598}} & 0.056 & \textbf{\textcolor{red}{5.270}} & \textbf{\textcolor{red}{6.657}} & 185.737 & 221.049 \\
        + FDS    & 0.644 & 11.493 & 13.603 & \textbf{\textcolor{red}{0.098}} & 6.682 & 9.013 & \textbf{\textcolor{red}{116.356}} & \textbf{\textcolor{red}{137.724}} \\
        + LDS \& FDS     & 0.682 & 10.947 & 12.766 & 0.021 & 5.416 & \textbf{\textcolor{red}{6.694}} & 502.807 & 586.384 \\
        + Ranksim      & 0.618 & 11.559 & 13.475 & 0.032 & 5.974 & 7.337 & 355.652 & 414.581 \\
        + Balanced MSE & 0.606 & 11.204 & 13.258 & 0.085 & 6.064 & 7.763 & 130.907 & 154.901 \\
        + \textbf{Ours}  & \textbf{\textcolor{red}{0.686}} & \textbf{\textcolor{red}{9.495}} & \textbf{\textcolor{red}{12.029}} & \textbf{\textcolor{red}{0.158}} & \textbf{\textcolor{red}{4.963}} & 6.851 & \textbf{\textcolor{red}{59.846}} & \textbf{\textcolor{red}{75.815}} \\
        \bottomrule
    \end{tabular}
    }
    \label{tab:performance_metrics}
\end{table*}

\subsection{Clinical validation}
\subsubsection{Baseline characteristics}
There are a total of 12,277 samples with a final predicted vascular age younger than their chronological age by more than 10 years, and 10,739 samples with a final predicted vascular age older than their chronological age by more than 10 years, leaving 77,000 samples with a difference between -10 and 10 years. The baseline characteristics of the three subgroups are presented in Table \ref{tab:baseline_characteristics}.

\begin{table}
\centering
\caption{Risk of different outcomes, summarized by the HRs and p-values across the three groups.}
\resizebox{0.49\textwidth}{!}{
\begin{tabular}{lcccc}
\toprule
\textbf{Outcomes} & \multicolumn{2}{c}{\textbf{\makecell{> 10 years younger vs. ± 10years}}} & \multicolumn{2}{c}{\textbf{\makecell{> 10 years older vs. ± 10 years}}} \\
\cmidrule(r){2-3} \cmidrule(r){4-5}
 & \textbf{HR (CI 95\%)} & \textbf{p value} & \textbf{HR (CI 95\%)} & \textbf{p value} \\
\midrule
\multicolumn{5}{l}{\textbf{Adjusted by age, sex, and ethnic background}} \\
Death & 0.90 (0.82-0.99) & 0.04 & 1.22 (1.11-1.35) & <0.005 \\
CHD & 0.77 (0.70-0.84) & <0.005 & 1.30 (1.19-1.42) & <0.005 \\
Stroke & 0.88 (0.64-1.21) & 0.42 & 1.96 (1.51-2.55) & <0.005 \\
HF & 0.89 (0.77-1.03) & 0.11 & 1.20 (1.04-1.39) & 0.01 \\
Hypertension & 0.64 (0.59-0.69) & <0.005 & 1.51 (1.42-1.61) & <0.005 \\
Diabetes & 0.70 (0.62-0.80) & <0.005 & 1.37 (1.23-1.53) & <0.005 \\
Atherosclerosis & 0.77 (0.43-1.37) & 0.37 & 1.96 (1.25-3.08) & <0.005 \\
\midrule
\multicolumn{5}{l}{\textbf{\makecell[l]{Adjusted by age, sex, ethnic background,\\ hypertension, and diabetes}}} \\
Death & 0.91 (0.83-1.01) & 0.07 & 1.19 (1.08-1.31) & <0.005 \\
CHD & 0.78 (0.71-0.86) & <0.005 & 1.25 (1.15-1.37) & <0.005 \\
Stroke & 0.91 (0.66-1.25) & 0.55 & 1.84 (1.41-2.39) & <0.005 \\
HF & 0.92 (0.79-1.06) & 0.24 & 1.14 (0.98-1.31) & 0.08 \\
Atherosclerosis & 0.79 (0.44-1.40) & 0.42 & 1.86 (1.18-2.92) & 0.01 \\
\midrule
\multicolumn{5}{l}{\textbf{\makecell[l]{Adjusted by age, sex, ethnic background,\\ hypertension, diabetes, BMI, and dyslipidemia}}} \\
Death & 0.91 (0.83-1.01) & 0.07 & 1.19 (1.08-1.31) & <0.005 \\
CHD & 0.78 (0.71-0.86) & <0.005 & 1.25 (1.15-1.36) & <0.005 \\
Stroke & 0.92 (0.66-1.26) & 0.59 & 1.82 (1.40-2.37) & <0.005 \\
HF & 0.90 (0.78-1.04) & 0.15 & 1.14 (0.98-1.31) & 0.08 \\
Atherosclerosis & 0.80 (0.45-1.42) & 0.44 & 1.84 (1.17-2.89) & 0.01 \\
\bottomrule
\end{tabular}}
\label{outcomes HR}
\end{table}

\begin{table*}
\centering
\caption{Baseline characteristics of the test set samples with PPG records}
\fontsize{9}{15}\selectfont
\begin{tabularx}{\textwidth}{
>{\raggedright\arraybackslash}p{4.7cm}
>{\centering\arraybackslash}X
>{\centering\arraybackslash}p{2cm}
>{\centering\arraybackslash}X
>{\centering\arraybackslash}p{1.5cm}
>{\centering\arraybackslash}p{1.5cm}}
\toprule
\textbf{Characteristics} & \textbf{\makecell{> 10 years younger\\(n=12,277)}} & \textbf{\makecell{± 10 years\\(n=77,000)}} & \textbf{\makecell{> 10 years older\\(n=10,739)}} & \textbf{p value} & \textbf{\makecell{p value\\for trend} }\\
\midrule
Ethnic background, white, n (\%) & 11055 (90.05) & 69213 (89.88) & 9367 (87.22) & 6.41e-11 & 2.15e-11 \\
Sex, male, n (\%) & 3949 (32.16) & 41845 (54.34) & 7359 (68.53) & < 1e-64 & < 1e-64 \\
Age (years), mean (s.d.) & 57.69 (9.07) & 59.25 (8.58) & 58.08 (7.93) & < 1e-64 & NA \\
BMI (kg/m\textsuperscript{2}), mean (s.d.) & 26.97 (4.63) & 27.27 (4.77) & 27.48 (4.79) & 6.39e-19 & NA \\
Diabetes, n (\%) & 667 (5.43) & 4530 (5.88) & 648 (6.03) & 0.096 & 0.046 \\
Hypertension, n (\%) & 3041 (24.77) & 22328 (29.00) & 3425 (31.89) & 4.79e-33 & 2.13e-33 \\
CHD, n (\%) & 561 (4.57) & 4516 (5.86) & 629 (5.86) & 5.19e-8 & 9.50e-6 \\
HF, n (\%) & 79 (0.64) & 468 (0.61) & 50 (0.47) & 0.155 & 0.090 \\
Stroke, n (\%) & 142 (1.16) & 1105 (1.44) & 166 (1.55) & 0.024 & 0.011 \\
Atherosclerosis, n (\%) & 10 (0.08) & 91 (0.12) & 17 (0.16) & 0.238 & 0.090 \\
\bottomrule
\end{tabularx}
\label{tab:baseline_characteristics}
\end{table*}

\subsubsection{Cardiovascular events analysis}
\begin{figure*}
\centering
\begin{minipage}[b]{0.33\linewidth}
  \centering
  \subfloat[]{\includegraphics[width=\linewidth]{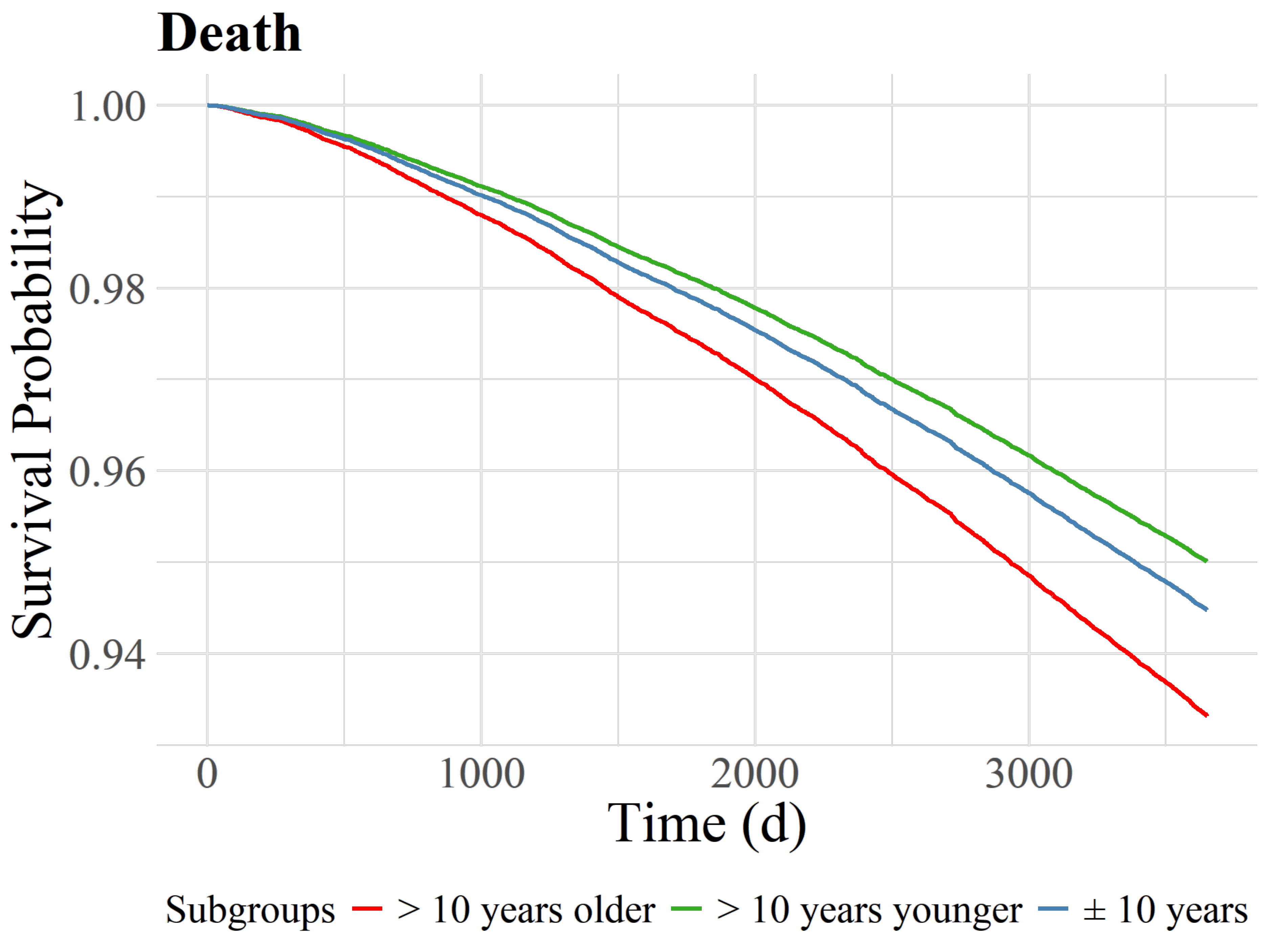}}
\end{minipage}
\hfill
\begin{minipage}[b]{0.33\linewidth}
  \centering
  \subfloat[]{\includegraphics[width=\linewidth]{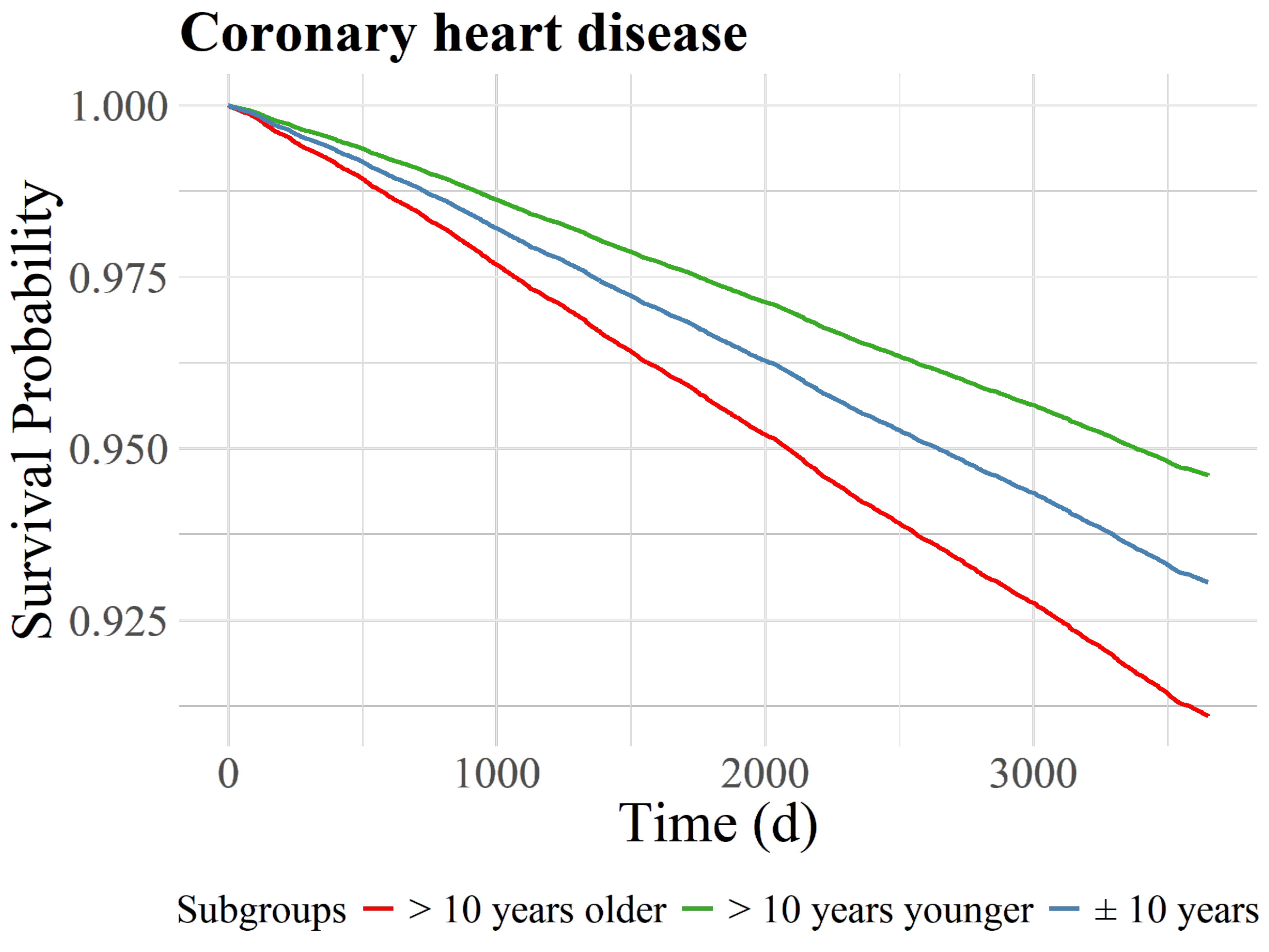}}
\end{minipage}
\hfill
\begin{minipage}[b]{0.33\linewidth}
  \centering
  \subfloat[]{\includegraphics[width=\linewidth]{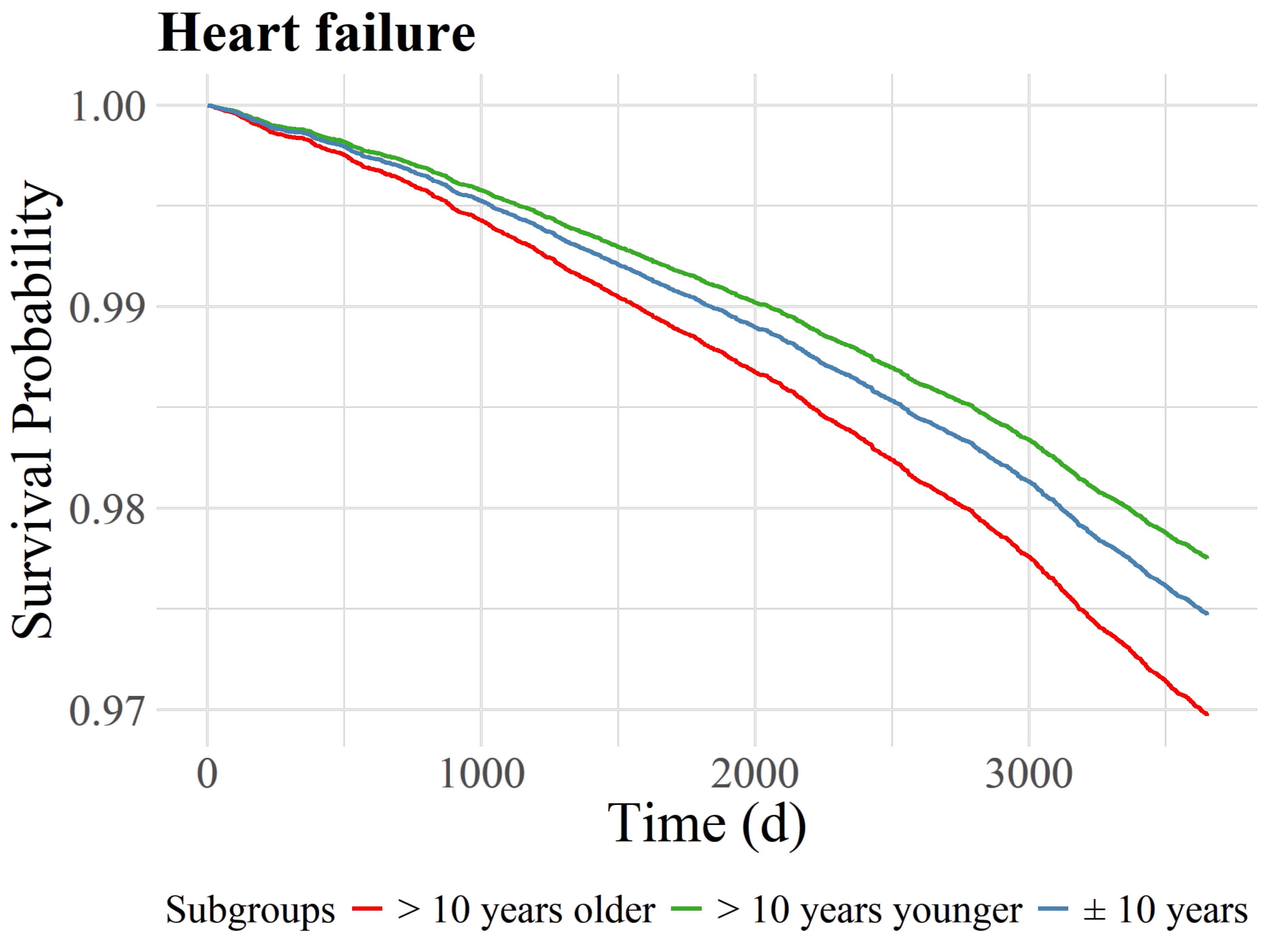}}
\end{minipage}

\caption{Adjusted KM curves for different outcomes, including death, CHD, and heart failure, by sex, age, and ethnic background.}
\label{fig:KM curves}
\end{figure*}

The results of the adjusted Kaplan-Meier (KM) curves are depicted in Figure \ref{fig:KM curves}. It can be observed that individuals with a predicted vascular age more than 10 years older than their chronological age have a higher risk for cardiovascular events, including death, CHD, and heart failure. Conversely, those with a predicted vascular age more than 10 years younger than their chronological age have a lower risk. Furthermore, the hazard ratios (HR) and 95\% confidence intervals (CI) for various cardiovascular events are summarized in Table \ref{outcomes HR}, revealing a consistent trend. These findings collectively underscore the potential of predicted vascular age as an indicator of cardiovascular health status.

\subsubsection{Analysis on arterial stiffness}
\begin{figure*}
\centering
\begin{minipage}[]{0.49\linewidth}
  \centering
  \subfloat[]{\includegraphics[width=\linewidth]{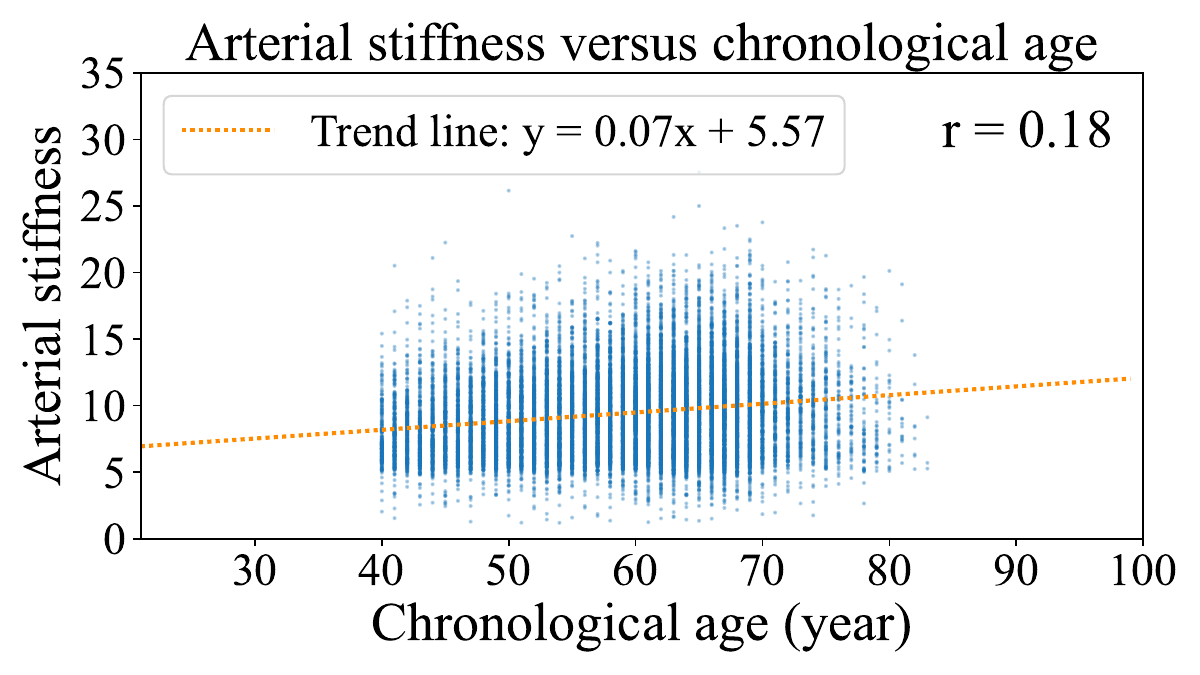}}
\end{minipage}
\hfill
\begin{minipage}[]{0.49\linewidth}
  \centering
  \subfloat[]{\includegraphics[width=\linewidth]{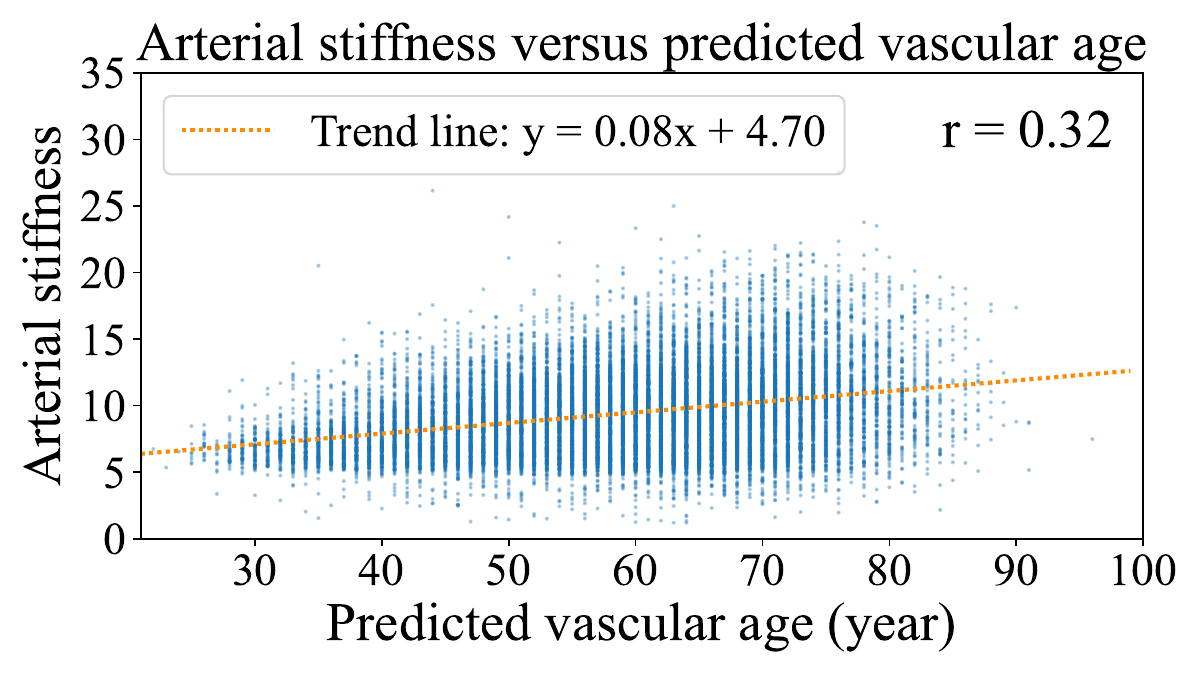}}
\end{minipage}
\caption{Scatter plots illustrating the relationship between arterial stiffness and chronological age (a) and predicted vascular age (b). $r$ denotes Pearson correlation coefficient.}
\label{fig: arterial stiffness}
\end{figure*}

Increased arterial stiffness is one of the most significant indicators of vascular aging. To verify the effectiveness of predicted vascular age, we examined its consistency with arterial stiffness. As shown in Figure \ref{fig: arterial stiffness}, a strong correlation exits between predicted vascular age and arterial stiffness. Furthermore, this correlation is stronger than that between chronological age and arterial stiffness, indicating that predicted vascular age more accurately represents vascular aging.

\section{Conclusions}
In conclusion, this study introduces a novel approach named the Dist loss to address the challenges of data imbalance in estimating vascular age from PPG signals. By incorporating the Dist loss, our deep learning model achieved state-of-the-art results on the extensive UK Biobank dataset, particularly in regions with small sample sizes. Furthermore, our analysis revealed a strong correlation between predicted vascular age and cardiovascular health, as evidenced by the relationship with various cardiovascular events over a 10-year follow-up period. These findings highlight the potential of predicted vascular age as a novel digital biomarker for assessing individual cardiovascular health status.

\section*{Acknowledgement}

Data from the UK Biobank, which is available after the approval of an application at \url{https://www.ukbiobank.ac.uk}. This work was supported by Beijing Natural Science Foundations (QY23040) and National Natural Science Foundation of China (No. 62102008).

\clearpage

\bibliographystyle{ACM-Reference-Format}
\balance
\bibliography{refs}

\end{document}